\pdfoutput=1

\documentclass[11pt]{article}

\usepackage[]{acl}

\usepackage{times}
\usepackage{latexsym}

\usepackage[T1]{fontenc}

\usepackage[utf8]{inputenc}

\usepackage{microtype}

\usepackage{graphicx}
\usepackage{amssymb,mathrsfs,amsmath, amsfonts}
\usepackage[ruled]{algorithm2e}

\title{Can depth-adaptive BERT perform better on binary classification task of Chinese texts}

\author{Jing Fan\textsuperscript{\rm 1*}, Xin Zhang\textsuperscript{\rm 2*}, Sheng Zhang\textsuperscript{\rm 3*}, Yan Pan\textsuperscript{\rm 4*}, Lixiang Guo\textsuperscript{\rm 5\dag}\\
    \textsuperscript{\rm *}National University of Defense Technology  \\
    \textsuperscript{\rm \dag}Information Engineering University  \\
    \textsuperscript{\rm 1}fanjing19@nudt.edu.cn, \\
    \textsuperscript{\rm 2}shinezhang\_nudt@qq.com,\\
    \textsuperscript{\rm 3}zhangsheng@nudt.edu.cn,\\
    \textsuperscript{\rm 4}panyan@mail.nwpu.edu.cn,\\
    \textsuperscript{\rm 5}zyzzyva\_glx@163.com
    }

\begin{document}
\maketitle
\begin{abstract}
In light of the success of transferring language models into NLP tasks and the magnitude of those language models, we ask whether the full language model is always the best and does it exist a simple but effective method to find the winning ticket in state-of-the-art deep neural networks without complex calculations. We construct a series of BERT-based models with different size by layer truncation and compare their predictions on 8 binary classification tasks. The results show there truly exist smaller sub-networks performing better than the full model. Then we present a further study and propose a simple method to shrink BERT appropriately before fine-tuning. Some extended experiments indicate that our method could save time and storage overhead extraordinarily both in fine-tuning and testing with little even no accuracy loss.
\end{abstract}

\section{Introduction}

Information retrieval is always of great concern, which

Recently, pretrain-then-finetune paradigm has led to significant improvements in a wide range of the NLP tasks. Some pre-trained language models, like BERT \cite{devlin2019bert}, achieve state-of-the-art performance on numerous tasks, such as sentiment analysis, acceptability analysis, semantic similarity analysis and  machine translation \cite{dolan2005automatically, warstadt2019neural, wang2018glue, yang2019enhancing, hoang2019aspect, yang2020towards}. For learning general knowledge as much as possible in pre-training stage, the language models are constructed with plenties of units (i.e., the total parameters in BERT\textsubscript{BASE} are 110M, and in BERT\textsubscript{LARGE} are 340M) \cite{devlin2019bert} . They are usually too huge for the downstream tasks. Taking BERT\textsubscript{BASE} as an example, fine-tuning it on some complex downstream tasks requires several hours to finish one epoch \cite{strubell2019energy}. That makes them impractical for resource-limited deployment scenarios and cost concerns incur, then hinders the applications. As we know, some compressed models (i.e., distilBERT, 66M \cite{sanh2019distilbert}) are capable of achieving comparable results. These lead us to think the magnitude of those large language models when transfer to target tasks, force us to compress models and accelerate their calculations \cite{mccarley2019structured}. However, existing model compression methods cost too much resources during fine-tuning, so we are eager to find a new way to solve the problem.

Existing model compression methods orienting to particular tasks fall into two categories, namely knowledge distillation \cite{bucilu2006model} and pruning \cite{frankle2018lottery}. For using knowledge distillation method, there needs to construct a complex teacher model and a much simple student. After the teacher fine-tuned thoroughly, the student imitates the teacher to predict and then makes decisions independently in test period. By doing so, resources like time and memory are saved in test period \cite{chia2019transformer, mirzadeh2020improved, yang2020textbrewer, wu2021one}. However, loading the teacher and student simultaneously and transferring knowledge from the teacher to the student during fine-tuning consume too much memory and time. Differently, pruning method only requires constructing one model, but mask parameters should be preserved to implement weight dropping. Matching the mask parameters with the original model and pruning the less important units during fine-tuning also need huge storage and time. \cite{joulin2016fasttext, liu2018rethinking, voita2019analyzing, gordon2019compressing, wang2020structured}. Some common issues could be concluded that (1) they all target at accelerating calculations in test stage at the cost of more resource-consuming during fine-tuning. And (2) all the downstream tasks apply the standard language models indiscriminately and treat them as the superior basis of pruning without taking the complexities of different tasks into consideration. Naturally, the matching relations between pre-trained language models and target tasks are neglected. Compressing models based on them is unreasonable.

According to the lottery ticket hypothesis \cite{frankle2018lottery, chen2020lottery}, there is some redundancy in over-parameterization BERT model. While some experiments about binary text classification tasks (like CoLA \cite{warstadt2019neural} and ChnSentiCorpHou \cite{tan2008empirical}) indicate that the standard BERT model is not always the best choice, and sometimes we could find sub-networks (winning tickets) surpassing the full BERT on some tasks via layer truncation (Details are provided in section \ref{sec2}). Motivated by this phenomenon, we propose to truncate BERT adaptively before fine-tuning and save resources both in fine-tuning and test stages with little cost. Compared with the conventional model compressing methods, our truncation method is valuable. Because fine-tuning time is much longer (hundreds times or more) than test time, while conventional methods fail to reduce it but need more in practice. A simple way to estimate the truncating position based on the characteristics of tasks is also provided. In addition, we conduct some experiments on 8 binary text classification tasks (Binary text classification tasks are simple enough to highlight the issues and are the basis of most complex NLP tasks). The results show that our method is simple but effective.

In summary, our contributions include:
\begin{itemize}
    \item We observe that full BERT is not always the best when fine-tuning on some particular downstream tasks. Sometimes shrinking BERT by simply truncating layers leads to capable even better performances with less overhead.
    
    \item We put forward a simple but effective truncation method to find out the winning ticket in BERT before fine-tuning, and provide a measure to estimate truncating position.
    
    \item Experiments in this paper show that our method could surpass the traditional compressing methods on 7 tasks and even predict more precisely than full BERT on 3 tasks. In addition, our method saves about 75\% fine-tuning time and 25\% storage than knowledge distillation and pruning methods on average. Also, 20\% storage and time are saved compared with original BERT-based models both in fine-tuning and testing while remaining near to 99\% accuracy on average.
\end{itemize}

\section{Full BERT is NOT always the best}
\label{sec2}
Knowledge distillation and pruning methods accelerate testing of models by devoting additional resources during fine-tuning. That is impractical for some edge devices. In order to save cost both in fine-tuning and testing, we propose to truncate several consecutive transformer layers from top of the BERT (12-layer) and preserve the bottom part as a new encoder. In this paper, we reduce BERT ranging from 0 to 11 layers and obtain 12 different encoders. Each encoder is appended a task-specific fully-connected layer and a sigmoid layer. All these 12 shrunk models (namely trans12 to trans1) are fine-tuned on eight text classification tasks, including CoLA \cite{warstadt2019neural} and ChnSentiCorpHou \cite{tan2008empirical}. More details of datasets, experimental settings and results are provided in Appendix \ref{apdx-a}, \ref{apdx-b} and \ref{apdx-c}.

\begin{figure*}
    \centering
    \includegraphics[width = 0.8\paperwidth] {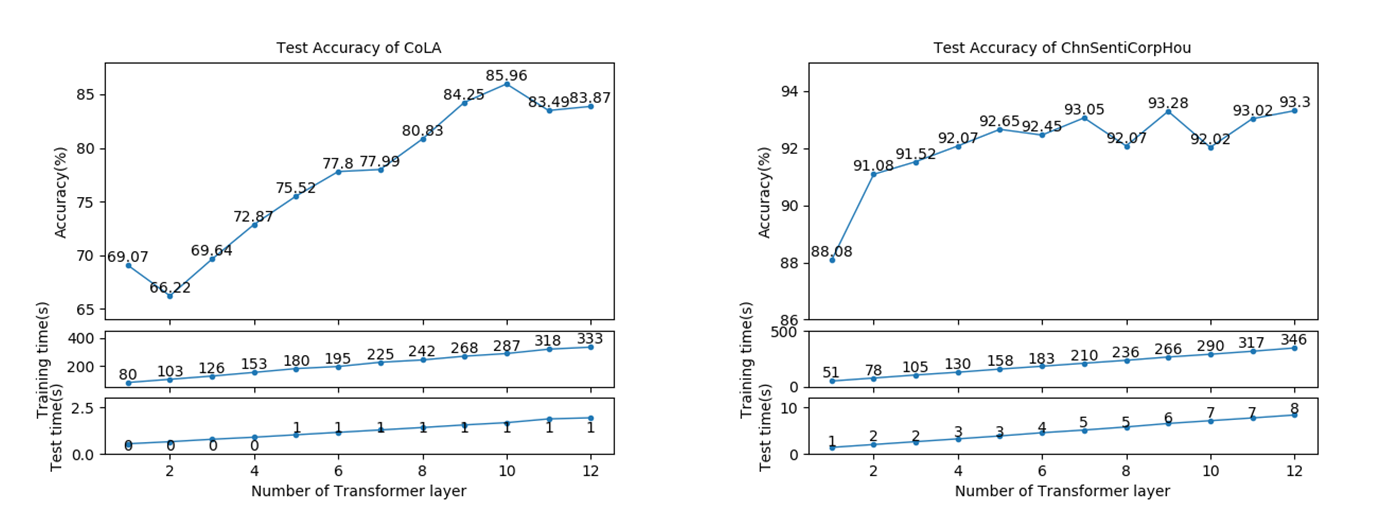}
    \caption{Accuracy, training time and test time of trans1-12. The reservation of accuracy arrives decimally hind two, and training and test time are rounded to the nearest whole number. In left subplot, trans5 reaches 90\% accuracy compared with trans12, and the accuracy of models with more than 8 layers becomes changing slightly. In the right, almost all these 12 models achieve more than 95\% accuracy of trans12, and the accuracy of models with more than 2 layers becomes fluctuating.}
    \label{fig1}
\end{figure*}

Here we take CoLA and ChnSentiCorpHou as examples and plot the test accuracy, training and test time of these 12 truncated models in Figure \ref{fig1}. There is no doubt that the training and test time are increasing dramatically as the number of transformer layer grows. However, we do not get far satisfying predictions with so many resources cost. With the increasement of the transformer layer, we find the accuracy is gradually fluctuating within a certain range. For CoLA (Figure \ref{fig1} (left)), we find two winning tickets surpassing the full BERT when number of transformer layers are 9 and 10. While the models with 8 and 11 layers are almost capable of predicting as precisely as full BERT-based model. As for ChnSentiCorpHou (Figure \ref{fig1} (right)), although we do not find a winning ticket better than full BERT, models with 2 to 12 layers predict similarly. It seems that smaller models could perform well enough with less time and memory compared with full BERT.

\section{Truncating BERT before fine-tuning}
\label{sec3}
The results in section \ref{sec2} show there exist truncated BERT models performing better than full BERT on some binary classification tasks. This inspires us that shrinking massive BERT for simple tasks is valuable and necessary. In order to clearly understand this phenomenon and find a way to solve problems well with less time and memory, we further take a closer look at these results and make some analyses about BERT truncation. In this section, we will provide some discussions about when, where and how to truncate BERT model.

\subsection{Where to truncate}

As discussed in section \ref{sec2}, the test accuracy becomes fluctuating when the number of transformer layers larger than some certain values. We then compare the predictions of each model and capture something interesting. When the test accuracies of two models are similar, the predictions on the same instance are also close. In addition, all these 12 models provide consistent predictions on most of the instances, and divisions only arise over a few indistinguishable ones. While these divisions would become less and less with the test accuracies being closer.

Specifically, given a dataset $\mathbb{S}$ containing $n$ examples $\rm{x_1},\rm{x_2},\cdots,\rm{x_{\textit{n}}}$. We construct 12 truncated BERT-based classification models, namely $M_1$ to $M_{12}$ (here, the subscript $i$ of $M$ denotes the number of transformer layers), to make predictions on the $\mathbb{S}$. The test accuracy of a model $M_i$ is denoted as ${acc}_i (i=1,\cdots,12)$. And the calculation of $M_i$ on $\rm{x_p} (p=1,\cdots,\textit{n})$ is formulated as:
\begin{eqnarray}
    M_{\textit{i}}(\rm{x_p})=\sigma(g_{\textit{i}}(h_{\textit{i}}(\rm{x_p}))),\\
    h_{\textit{i}}(\rm{x_p}) = f_{\textit{i}}(f_{\textit{i}-1}(\cdots(f_1(\rm{x_p}))\cdots)).
\end{eqnarray}
Where $\sigma$ is the sigmoid function, $g_i (\cdot)$ is the function of a fully-connected layer, $h_i(\cdot)$ is the function of the encoder, and $f_i (\cdot)$ is the function of the $i$-th transformer encoder layer.
As mentioned above, when $acc_i$ is close to $acc_j$, we find $M_i(\rm{x_p})$ is also close to $M_j(\rm{x_p})(p=1,\cdots,\textit{n};\textit{i},\textit{j}=1,\cdots,12,\ and\ i<j)$. For classification task, we know $\sigma(g_i(\cdot))$ is a generalized linear transformation targeting at finding a hyperplane separating examples best without transforming the representations. Naturally, the prediction $M_i(\cdot)$ mainly depends on the transformation of the encoder $h_i(\cdot)$. The higher separability of the hidden states produced by the encoder is, the higher prediction probability $M_i(\cdot)$ and the higher test accuracy $acc_i$ of the model are. As a result, we could use the hidden states to help us decide where to truncate.

Inspired by the distribution of points in high dimensional space, we treat the output representations of transformer layers in encoder as coordinates of point, and apply the separability theory of point set to expressing the distribution of examples in our datasets. We truncate BERT targeting at producing representations of dataset with maximum seperability.

Here, we introduce three widely used separability measures, namely Class Scatter Matrices (CSM) \cite{duda1973pattern}, Thornton’s Separability Index (SI) \cite{thornton2002truth}, and Hypothesis Margin (HM) \cite{gilad2004margin}.

\textbf{Class Scatter Matrices, CSM} : is a fraction of trace of between-cluster scatter and within-cluster scatter. It evaluates the overall distribution of an entire dataset.

\begin{eqnarray}
    \label{eq3}
    CSM=tr(S_B )/tr(S_W ),\\
    S_B = \sum_{i=1}^{c} n_i(\rm{m}_\textit{i}-\rm{m})(\rm{m}_\textit{i}-\rm{m})^T,\\
    S_W = \sum_{i=1}^{c} \sum_{j=1}^{n_i} (\rm{x}_{\textit{i},\textit{j}}-\rm{m}_\textit{i})(\rm{x}_{\textit{i},\textit{j}}-\rm{m}_\textit{i})^T.
\end{eqnarray}
Where $tr(\cdot)$ denotes the trace operation: calculating the sum of one’s diagonal elements. $c$ and $n_i$ are the number of classes in a dataset and number of examples in class $i$ respectively. While $m$ and $m_i$ are the mean vectors of whole dataset and class $i$ separately. $\rm{x_{\textit{i},\textit{j}}}$ is the $j$-th instance in class $i$. The within-cluster scatter and between-cluster scatter are denoted as $S_W$ and $S_B$.

\textbf{Separability Index, SI}: calculates the average number of instances that share the same category label as their nearest neighbors. It is a measure evaluating the class overlap.

\begin{equation}
    SI = \frac{\sum_{i=1}^{n} (\phi(\rm{x_i}) +\phi(\rm{x}_i^{\prime})+1) mod\ 2}{n},
    \label{eq4}
\end{equation}
Where $\phi(\rm{x_\textit{i}})(\phi(\rm{x_\textit{i}} )\in \{0,1\})$ is the label of $\rm{x_\textit{i}}$, $\rm{x}_\textit{i}^{\prime}$ is the nearest neighbor of $\rm{x}_\textit{i}$, and $n$ is the number of examples in a dataset.

\textbf{Hypothesis Margin, HM}: measures the sum of the distance between the hypothesis and the closest hypothesis that assigns the alternative label to the given instance.

\begin{equation}
    \begin{aligned}
        HM = \sum_{i=1}^{n} \frac{1}{2} (\| \rm{x}_\textit{i} - nearmiss(\rm{x}_\textit{i})\| - \\
        \| \rm{x}_\textit{i} - nearhit(\rm{x}_\textit{i})\| ),
    \end{aligned}
    \label{eq5}
\end{equation}
Where nearhit($\rm{x}_\textit{i}$) and nearmiss($\rm{x}_\textit{i}$) denote the nearest instance to $\rm{x}_\textit{i}$ with the same and different label, respectively. And $n$ is the number of instances in a dataset. $\| \cdot \|$ denotes the Euclidean norm.

According to the definitions above mentioned, the larger these three separability values are, the more easily the examples are classified. We could apply our truncation method with the assistance of the separability measures.

\subsection{When to truncate}

As we all known, BERT is pre-trained on two general tasks, namely masked language model (MLM) and next sentence prediction (NSP). It means there is little disturbance of wrong prediction corresponding to specific task brought in. Making decisions on the basis of original pre-trained BERT is more objective.

What’s more, once the fine-tuning processing starts, the predicting error would radially propagate back to each layer of the entire network. Breaking the network during this stage leads to reconstruct new matching relations among whole net, which is far more time-consuming.

On account of these factors, we decide to truncate BERT before fine-tuning. If we can make a reliable decision about where to truncate before fine-tuning, resources like time and storage would be saved extraordinarily.

\subsection{How to truncate}

In this sub-section, we provide a general algorithm about truncating BERT (12-layer) properly for downstream tasks by means of   separability measures (Algorithm \ref{alg1}).

Given a target binary classification task and the corresponding dataset, we can estimate where to truncate before fine-tuning without building up the network. Our three-step method is simple enough and time-saving. In particular, all the instances in the dataset should be vectorized with BERT embeddings in the first step. Secondly, the vectorized instances are sent into the pre-trained BERT model and propagate forward once without backward propagation. In this step, we will obtain 12 hidden representations from different transformer layers of the BERT for every instance. The hidden representations produced by the same transformer layer are packed into a common set. Then in the third step, we calculate separability values for these 12 representation sets. The number of transformer layers related to the maximum separability is the suggested number of layers in the truncated BERT. We truncate BERT according to that and append the task-specific layer on to form the final model.

\begin{algorithm}[h]
 \LinesNumbered
 \KwData{$\mathbb{S} = \{x_1,x_2,\cdots,x_n \}$}
 \For{$x_i(i=1,\cdots,n)$ in $\mathbb{S}$}{
 Vectorize $x_i$ as $\textrm{x}_i$\;
 Send $\textrm{x}_i$ in BERT\; 
 \For{each transformer layer $l_j(j=1,\cdots,12)$ in BERT}
 {
  Calculate hidden state $\textrm{h}_{\textit{i},\textit{j}}$ \;
  Collect $\textrm{h}_{\textit{i},\textit{j}}$ into $v_j (v_j=\{\textrm{h}_{1,\textit{j}},\cdots,\textrm{h}_{\textit{n},\textit{j}}\})$\;
 }
 }
 \For{each $v_j(j=1,\cdots,12)$} {
 Calculate separability value ($\text{csm}_j$, $\text{si}_j$ or $\text{hm}_j$)\;
 }
 Truncate BERT according to the maximum separability value
 \caption{Truncating algorithm}
 \label{alg1}
\end{algorithm}

In this way, we can truncate BERT adaptively before fine-tuning according to target tasks. Putting these truncated BERT into force could save as much as time and memory with little expense.

\section{Experiments}
\label{sec4}
Observations in section \ref{sec2} inspires us that the full BERT is not always the best choice for some binary classification tasks. And in section \ref{sec3} we provide a simple but clear analysis about this phenomenon, and introduce an effective method to shrink BERT for downstream tasks. In this section we will apply our method to 8 binary classification tasks and give a further exploration.

\begin{table*}[htbp]
    \centering
    \resizebox{\textwidth}{!}{
        \begin{tabular} {cccccc}
        \hline
        \textbf{Corpus} & \textbf{Size} & \textbf{ Input Style} & \textbf{Task} & \textbf{Language} & \textbf{Degree of Difficulty}\\
        \hline
        ChnSentiCorpHou & 6k & single sentence & sentiment classification & Chinese & $\bigstar$\\
        WaiMai & 12k & single sentence & sentiment classification & Chinese & $\bigstar \bigstar$ \\
        SST-2 & 67.8k & single sentence & sentiment classification & English & $\bigstar \bigstar \bigstar$\\ 
        CoLA & 9k & single sentence & acceptability analysis & English &  $\bigstar \bigstar \bigstar \bigstar$\\ 
        Nikon-JD & 2.6k & sentence pair & aspect-level sentiment classification & Chinese & $\bigstar \bigstar$\\ 
        AFQMC & 38.6k & sentence pair & semantic similarity & Chinese & $\bigstar \bigstar \bigstar$\\ 
        MRPC & 5.8k & sentence pair & semantic similarity & English & $\bigstar \bigstar \bigstar$\\ 
        RTE & 2.7k & sentence pair & natural language inference & English & $\bigstar \bigstar \bigstar \bigstar \bigstar$\\ 
        \hline
        \end{tabular}
    }
    \caption{Corpora information. All these 8 are binary text classification tasks with a single sentence or sentence pair as the input. In the last column, the more the stars are, the more difficult the task is.}
    \label{table1}
\end{table*}

\subsection{Datasets and tasks}
In this sub-section, we will give a brief introduction about 8 corpora used in this paper, including ChnSentiCorpHou, WaiMai, SST-2, CoLA, Nikon-JD, AFQMC, MRPC and RTE. The ChnSentiCorpHou is a sub-set of a series of Chinese sentiment analysis corpora organized by Tan et al. \cite{tan2008empirical}. The WaiMai is a binary classification corpus about reviews of delivery service and takeaway food\footnote{https://github.com/SophonPlus/ChineseNlpCorpus/tree/master/datasets/waimai\_10k}. The SST-2 \cite{socher2013recursive}, CoLA \cite{warstadt2019neural}, MRPC \cite{dolan2005automatically} and RTE \cite{dagan2005pascal, haim2006second, giampiccolo2007third, bentivogli2009fifth} come from the general language understanding evaluation (GLUE) benchmark \cite{wang2018glue}, and AFQMC\footnote{https://dc.cloud.alipay.com/indexn\#/topic/intro?id=3}  is selected from the Chinese language understanding evaluation (CLUE) benchmark \cite{xu2020clue}. The Nikon-JD corpus is organized by our team. More details of these datasets are provided in Appendix \ref{apdx-a}. In Table \ref{table1}, we list some critical information about these corpora. We select two Chinese and two English single sentence classification tasks. In addition, two Chinese and two English sentence pair classification tasks are also chosen. The assignments in this section vary from sentiment classification and acceptability analysis to semantic similarity and natural language inference. Different assignment is related to different size of dataset and difficulty degree.

\subsection{Implementations}
We select the 12-layer BERT model (BERT\textsubscript{BASE}) as our basic language model and apply the truncation method on these 8 datasets. For each single sentence task, we put a [CLS] and [SEP] at the beginning and the end of the given text separately. And for sentence pair task, we also concatenate two texts into a sequence with [SEP] token being the separator, besides the special pre-processing at beginning and end. Then send processed sequences into the BERT for following procedures.

Before constructing models and fine-tuning, we utilize whole dev sets of ChnSentiCorpHou, Nikon-JD, MRPC and RTE, and randomly sample about 50\%, 50\%, 10\% and 10\% examples from dev sets of SST-2, CoLA, WaiMai and AFQMC separately. Then we calculate CSM, SI and HM with these prepared instances for each task, and according to that truncate BERT model adaptively. Next, we construct classification models based on the truncated BERT models, then fine-tune and test them on above mentioned corpora. More experimental details are in Appendix \ref{apdx-b}.

For comparison, we also set up some baselines:

\textbf{BERT base}: We construct binary classification models based on BERT\textsubscript{BASE} \cite{devlin2019bert} as the standard models. 

\textbf{Distil-Pretrain}: We build up classification models based on the distilled language models (DistilBERT \cite{sanh2019distilbert} and albert-chinese-tiny\footnote{https://huggingface.co/clue/albert\_chinese\_tiny}). In this way, the knowledge distillation theory is used in pre-training period.

\textbf{Distil-task-biLSTM}: We fine-tune a BERT-based classification model as the teacher, and construct a student using the biLSTM as the encoder with the assistance of the attention mechanism. Then let the student imitates prediction of the teacher. 

\textbf{Distil-task-kimCNN}: We construct a student according to the kimCNN \cite{2014Convolutional}. While the teacher models are same as those of Distil-task-biLSTM.


\textbf{Pruning-AGP}: We use the Automated Gradual Pruner (AGP) \cite{zhu2017prune} to implement the task-specific magnitude pruning on our tasks. Here we refer to the implementation provided in the Distiller library \cite{zmora2019neural}.

\textbf{Random truncation}: Random truncation means selecting a random number ranging from 0 to 11 as the number of layers that should be truncated. Here, we use the mean test accuracy of 12 various truncated BERT-based models to denote the random accuracy.

Note that all these baselines use the BERT embeddings, the batch size and max sequence length are same as those of our models (Table \ref{table6}). Differently, during implementing task-specific distillation, we set the learning rate as 1e-3.

\subsection{Results}
In order to evaluate whether our separability measures are appropriate or not and then choose a better one for truncating BERT, we firstly plot three measures with the test accuracies in Figure \ref{fig2} (Here we only provide the results of CoLA and ChnSentiCorpHou, while others are shown in Appendix \ref{apdx-d}). Obviously, we could easily find that the tendencies of these three measures are similar to the test accuracies to some extent. As revealed in the Equation \ref{eq3}-\ref{eq5}, distance-based 
measures (CSM and HM) are more exact than the one based on number of instances (SI), while HM is easily influenced by some specific examples and loses effectiveness. Among these three measures, the CSM is more stable. To make a quantitative analysis, we use the Pearson’s correlation coefficient (PCC) to evaluate the relations. The average of PCC in Table \ref{table2} also indicates that the CSM is more relevant to test accuracy. As a result, we choose the CSM measure to help us truncate BERT.

\begin{table}[]
    \centering
    \begin{tabular}{cccc}
    \hline
     & CSM & SI & HM\\
    \hline
    ChnSentiCorpHou & 0.7942 & 0.8204 & 0.4208\\
    WaiMai & 0.2171 & 0.1927 & -0.6054 \\
    SST-2 & 0.8697 & -0.1703 & 0.5242\\
    CoLA & 0.9655 & 0.5444 & 0.8269\\
    Nikon-JD & 0.9517 & 0.8014 & 0.7963\\
    AFQMC & 0.7709 & -0.2119 & 0.7073\\
    MRPC & 0.2544 & 0.7389 & 0.4105\\
    RTE & 0.8720 & 0.1090 & -0.2889\\
    Avg & 0.7119 & 0.3530 & 0.3489\\
    \hline
    \end{tabular}
    \caption{Pearson’s correlation coefficient of CSM, SI and HM. The items in the last row are the average values of the corresponding measures.}
    \label{table2}
\end{table}

\begin{figure*}
    \centering
    \includegraphics[width = 0.8\paperwidth]{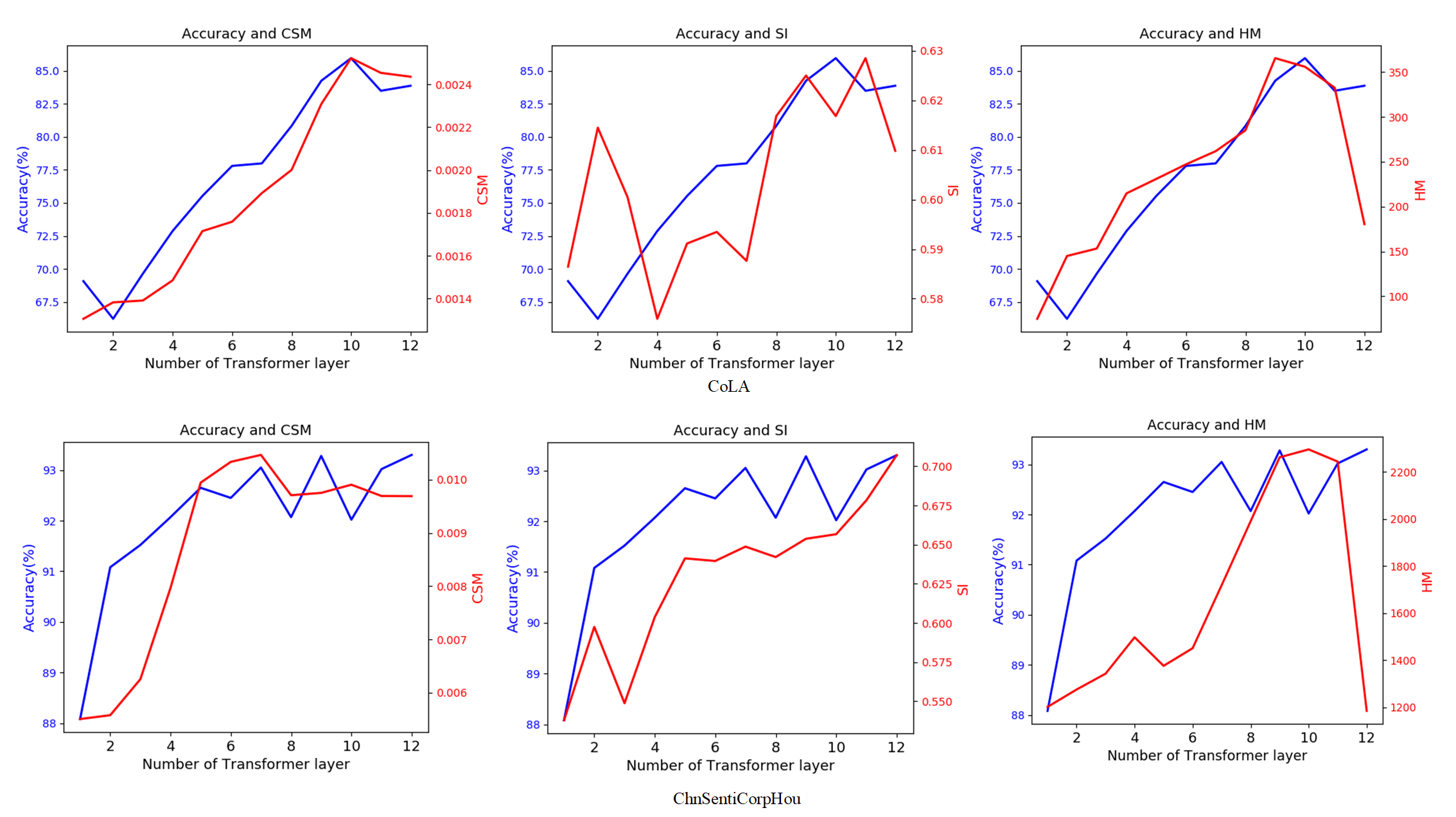}
    \caption{CSM, SI and HM Curves of CoLA and ChnSentiCorpHou. The test accuracy is shown with a royal blue line, and the separability value is drawn with a red line.}
    \label{fig2}
\end{figure*}

As shown in Figure \ref{fig2} and Appendix \ref{apdx-d}, when the number of transformer layers are 7, 11, 12, 10, 11, 12, 5 and 9, the CSM values reach the maximum on ChnSentiCorpHou, WaiMai, SST-2, CoLA, Nikon-JD, AFQMC, MRPC and RTE, respectively. So we accept these suggestions and truncate BERT for these tasks. In this way, the test accuracies of our truncated model are 93.05\%, 90.83\%, 92.26\%, 85.96\%, 97.31\%, 74.12\%, 76.52\% and 63.68\%.

The test accuracy, processing time and scale of parameters are listed in Table \ref{table3}, Table \ref{table4} and Table \ref{table5}, respectively. Obviously, our method, Truncated-CSM, is capable of the 99\% language understanding capabilities of the full model in average, and almost surpasses the other compressing methods on all these tasks except the MRPC. Also, about 20\% of time and memory are saved both in fine-tuning and test with little additional cost, which is impossible for knowledge distillation and pruning. What’s more, our method is simple to implement and transfer to other tasks. On the contrary, we can draw conclusions from Table \ref{table4} and Table \ref{table5} that distillation methods are too complex to apply and consume far more time and storage from building up the teacher model to obtaining the student. The pruning methods are of the same circumstances in fine-tuning. The overhead of knowledge distillation and pruning methods in fine-tuning is awful and unbearable. However, we do not get satisfying predictions with these two kinds of ways.

\begin{table*}[]
    \centering
    \resizebox{\textwidth}{!}{
        \begin{tabular}{ccccccccc}
        \hline
        Model & ChnSentiCorpHou (\%) & WaiMai (\%) & SST-2 (\%) & CoLA (\%) & Nikon-JD (\%) & AFQMC (\%) & MRPC (\%) & RTE (\%)\\
        \hline
        BERT base & 93.30 & 90.12 & 92.26 & 83.87 & 97.01 & 74.12 & 83.59 & 65.81\\
        Distil-Pretrain & 91.17 & 89.82 & 91.71 & 81.40 & 90.72 & 51.88 & 82.09 & 59.56\\
        Distil-task-biLSTM & 87.50 & 89.57 & 87.10 & 68.31 & 89.32 & 64.18 & 64.17 & 50.88\\
        Distil-task-kimCNN & 86.48 & 88.45 & 85.23 & 70.59 & 91.29 & 63.51 & 67.59 & 52.08\\
        Pruning-AGP & 92.65 & 87.11 & 91.32 & 82.73 & 95.23 & 69.95 & 77.33 & 57.50\\	
        Random truncation & 92.05 & 90.21 & 90.43 & 77.29 & 95.66 & 70.90 & 76.68 & 64.21\\
        Truncated-CSM & \textbf{93.05} & \textbf{90.83} & \textbf{92.26} & \textbf{85.96} & \textbf{97.31} & \textbf{74.12} & 76.52 & \textbf{63.68}\\
         \hline
        \end{tabular}
    }
    \caption{Test accuracy of 8 tasks in this paper (including 2 tasks in section \ref{sec2}). Bond font in the last row indicates the test accuracy of our truncated model is larger than other model compressing methods (except BERT base).}
    \label{table3}
\end{table*}

\begin{table*}[]
    \centering
    \resizebox{\textwidth}{!}{
        \newcommand{\tabincell}[2]{\begin{tabular}{@{}#1@{}}#2\end{tabular}}
        \begin{tabular}{ccccccccc}
        \hline
        Model & ChnSentiCorpHou & WaiMai & SST-2 & CoLA & Nikon-JD & AFQMC & MRPC & RTE\\
        \hline
        BERT base & 347 & 439 & 3918 & 334 & 104 & 1617 & 159 & 177\\
        Distil-Pretrain & - & - & - & - & - & - & - & -\\
        Distil-task-biLSTM & \tabincell{c}{347+841} & \tabincell{c}{439+613} & \tabincell{c}{3918+5493} & \tabincell{c}{334+761} & \tabincell{c}{104+248} & \tabincell{c}{1617+3933} & \tabincell{c}{159+371} & \tabincell{c}{177+284}\\
        Distil-task-kimCNN & \tabincell{c}{347+738} & \tabincell{c}{439+570} & \tabincell{c}{3918+5052} & \tabincell{c}{334+1312} & \tabincell{c}{104+455} & \tabincell{c}{1617+3836} & \tabincell{c}{159+1113} & \tabincell{c}{177+261} \\
        Pruning-AGP & 10516 & 5175 & 21389 & 2326 & 1970 & 16643 & 2467 & 2004\\	
        Random truncation & - & - & - & - & - & - & - & -\\
        Truncated-CSM & 211 & 401 & 3918 & 288 & 96 & 1617 & 83 & 134\\
         \hline
        \end{tabular}
    }
    \caption{Processing time of 8 tasks in this paper (including 2 tasks in section \ref{sec2}). For distillation-based method, it includes fine-tuning time of teacher model and distillation time of student model. While for truncation-based method, it includes calculation time of separability measure and fine-tuning time. The unit of measurement is second (s). Because the total time of Distil-Pretrain and Random truncation are not available, here we do not provide related time.}
    \label{table4}
\end{table*}

\begin{table*}[]
    \centering
    \resizebox{\textwidth}{!}{
        \newcommand{\tabincell}[2]{\begin{tabular}{@{}#1@{}}#2\end{tabular}}
        \begin{tabular}{ccccccccc}
        \hline
        Model & ChnSentiCorpHou & WaiMai & SST-2 & CoLA & Nikon-JD & AFQMC & MRPC & RTE\\
        \hline
        BERT base & 102 & 102 & 109 & 109 & 102 & 102 & 109 & 109\\
        Distil-Pretrain & - & - & - & - & - & - & - & - \\
        Distil-task-biLSTM & \tabincell{c}{102+9} & \tabincell{c}{102+9} & \tabincell{c}{109+9} & \tabincell{c}{109+9} & \tabincell{c}{102+9} & \tabincell{c}{102+9} & \tabincell{c}{109+9} & \tabincell{c}{109+9}\\
        Distil-task-kimCNN & \tabincell{c}{102+7} & \tabincell{c}{102+7} & \tabincell{c}{109+7} & \tabincell{c}{109+7} & \tabincell{c}{102+7} & \tabincell{c}{102+7} & \tabincell{c}{109+7} & \tabincell{c}{109+7}\\
        Pruning-AGP & 102 & 102 & 109 & 109 & 102 & 102 & 109 & 109\\	
        Random truncation & - & - & - & - & - & - & - & -\\
        Truncated-CSM & 67 & 95 & 109 & 95 & 95 & 102 & 53 & 88\\
         \hline
        \end{tabular}
    }
    \caption{Scale of parameters of 8 tasks in this paper (including 2 tasks in section \ref{sec2}). All values are scaled by 1M.}
    \label{table5}
\end{table*}

According to these results, we confirm our method is effective and provide the following suggestions:
\begin{itemize}
    \item Only 1 layer of transformer encoder is enough to perform well on most simple tasks.
    \item For some complex tasks, a few more layers (3-5 layers) could produce satisfying results.
    \item More exact estimation could be made by randomly sampling some instances to calculate CSM value.
\end{itemize}

\section{Related Work}

In this section, we briefly review some work about pre-trained language models, knowledge distillation and model pruning.

\textbf{Pre-trained language models}: Since about 2017, there gradually appear some complex language models. The widely used BERT proposed by Devlin et al. \cite{devlin2019bert} achieves pretty good results on GLUE \cite{wang2018glue}, SQuAD \cite{rajpurkar2016squad} and SWAG \cite{zellers2018swag}. Before long, XLNet \cite{yang2019xlnet}, RoBERTa \cite{liu2019roberta} and some others \cite{zafrir2019q8bert, shen2020q, wang2021kepler} emerge and push the evaluation scores of NLP tasks much higher. With the development of pre-trained language models, a wide range of NLP tasks obtain new state-of-the-art results again and again, including question answering \cite{mccarley2019structured}, language inference \cite{levesque2012winograd, williams2018broad}, text classification \cite{joulin2016fasttext} and text summarization \cite{barzilay1999using}. Consequently, the language models has been an indispensable part in solving NLP assignments.

\textbf{Knowledge distillation}: Knowledge distillation is a method that can transfer knowledge from an ensemble or a large highly regularized model into a much smaller, distilled model. This concept is proposed by Caruana et al. \cite{bucilu2006model} and further extended to Neural Networks by Hinton’s team \cite{hinton2015distilling}. Some like SANH et al. \cite{sanh2019distilbert} and Jiao et al. \cite{jiao2020tinybert} only use this theory to compress models and accelerate processing. More are inspired not only to reduce the models but also transfer extra knowledge by means of changing input style or other factors during distillation \cite{chen2019distilling, ma2020adversarial, chen2020online, wang2020structure, sun2020knowledge}.

\textbf{Model pruning}: Existing model pruning work could classify into two categories, namely structured pruning and unstructured pruning. The structured pruning means removing models by dropping several entire layers or some coherent groups of blocks \cite{michel2019sixteen, fan2019reducing, wang2020structured}. In contrast, unstructured pruning is a form of dropout that compresses model by reducing weights independently \cite{frankle2018lottery, gordon2019compressing, prasanna2020bert, chen2020lottery}.

\section{Conclusion}
In this paper, we conduct a series of experiments to exploring whether it exist winning tickets when transferring the language models in downstream binary classification tasks and how to find them conveniently and effectively. Different from conventional model compressing methods, we insist to truncate BERT by a brute force mean that is dropping out several consecutive transformer layers once. In addition, a simple way to estimate the size of BERT before fine-tuning is provided by us. In this way, the resources are saved tremendously without losing too much accuracy. More effectively, some simple task could be solved only use one transformer layer sometimes.

Currently, there is a little work studying the over-parameterization problem of state-of-the-art networks based on language models. And researchers still fail to comprehend and explain these complex and massive deep neural networks. Also, there is little work solving the resource-consuming problem well. However, these problems should be attached importance to and solved well before the development of neural models reaching a higher level. Now we throw away a brick in order to attract more researchers to pay attention to then solve these problems.

In future, we will extend these experiments to other more complex NLP tasks and other language models, and explore more effective ways to find winning tickets in language models. Also, we are eager to find out more effective ways to solve these problems better.


\bibliography{anthology,custom}
\bibliographystyle{acl_natbib}

\newpage
\appendix

\section{Datasets Information}
\label{apdx-a}

Here, we provide more exact descriptions about corpora used in this paper.

\textbf{ChnSentiCorpHou}: The ChnSentiCorpHou is a Chinese sentiment corpus about hotel comments collected by Tan et al. \cite{tan2008empirical}. It consists of 3000 positive and 3000 negative reviews about hotel. We conduct our single sentence classification task on it. When given a comment, we should predict the corresponding polarity.


\textbf{WaiMai}: The WaiMai is a Chinese sentiment corpus about comments of delivery service and takeaway food\footnote{https://github.com/SophonPlus/ChineseNlpCorpus/tree/master/datasets/waimai\_10k}. It consists of about 4000 positive and 8000 negative examples.

\textbf{SST-2}: The Stanford Sentiment Treebank \cite{socher2013recursive} consists of about 70k English sentences from movie reviews. These sentences are annotated with positive or negative polarity. The target of this task is to predict the sentiment category when given a sentence.

\textbf{CoLA}: The Corpus of Linguistic Acceptability \cite{warstadt2019neural}  consists of a plenty of English acceptability judgments drawn from books and journal articles on linguistic theory. There are 10k examples in this corpus, and each of which is a sentence annotated with a binary label indicating whether it is a grammatical English sequence or not. Note that for privacy limitation, we only utilize the training and dev set.

\textbf{Nikon-JD}: The Nikon-JD is a Chinese aspect-level sentiment classification dataset consisting of 2.6k product reviews crawled from E-commerce platform, namely JingDong\footnote{https://www.jd.com/}. We organize 3 professors annotating these reviews. Given a review and a target aspect category, all those 3 annotators should make a decision whether the review expresses a positive polarity or not toward the aspect. Then we label it in terms of the majority.

\textbf{AFQMC}: The Ant Financial Question Matching Corpus \cite{xu2020clue} comes from the Ant Technology Exploration Conference Developer competition. There are 42.5k instances in this corpus, each of which is a sentence pair. It is a binary classification task that targets at predicting whether a pair of sentences are semantically similar.

\textbf{MRPC}: The Microsoft Research Paraphrase Corpus \cite{dolan2005automatically} is a set of English sentence pairs automatically extracted from online news platforms, each pair annotated with a label indicating whether the sentences in a pair are semantically equivalent. There are approximately 5.8k pairs in this corpus.

\textbf{RTE}: The Recognizing Textual Entailment (RTE) \cite{wang2018glue} dataset consists of 2.7k sentence pairs. Each pair of sentences is labeled with a binary category signifying whether one sentence could entail the other or not.

\section{Experimental Settings}
\label{apdx-b}

Full ChnSentiCorpHou, WaiMai, SST-2 and MRPC corpora are obtained from the original sources . Due to the privacy limitation, we only get the training and dev set of CoLA, AFQMC and RTE. For CoLA and AFQMC, we treat the original dev sets as our new test sets and randomly sample 10\% instances from the original training sets forming the new dev sets while the remaining examples forming the new training sets. Then we fine-tune classification models on the training and dev sets and implement test on the test sets for WaiMai, SST-2, CoLA, AFQMC and MRPC. We perform 5-fold cross-validations on the remaining three datasets, namely ChnSentiCorpHou, Nikon-JD and RTE, which means partitioning the examples into 5 parts randomly, choosing a different fold as the test set with accuracy calculated on, and leaving the remaining 4 folds to form the training and test set each time. In the evaluation step, we use the average accuracy of 5 test sets as the final result.

\begin{table}[htbp]
    \centering
    \newcommand{\tabincell}[2]{\begin{tabular}{@{}#1@{}}#2\end{tabular}}
    \resizebox{0.5\textwidth}{!}{
        \begin{tabular}{ccccc}
            \hline
            Corpus & Batch Size & Learning rate & Max Sequence length & Device \\
            \hline
            ChnSentiCorpHou & 32 & 1e-5 & 512 & RTX3090$\times$2 \\
            WaiMai & 32 & 1e-5 & 256 & RTX3090$\times$2 \\
            SST-2 & 32 & 2e-5 & 128 & V100$\times$1 \\
            CoLA & 32 & 2e-5 & 128 & RTX3090$\times$2 \\
            Nikon-JD & 32 & 1e-5 & 256 & RTX3090$\times$2 \\
            AFQMC & 32 & 5e-6 & 256 & RTX3090$\times$2 \\
            MRPC & 32 & 1e-5 & 128 & RTX3090$\times$2 \\
            RTE & 32 & 1e-5 & 256 & V100$\times$1 \\
            \hline
    \end{tabular}
    }
    \caption{Experimental settings.}
    \label{table6}
\end{table}

We use 12-layer Chinese BERT-wwm-ext\footnote{https://huggingface.co/hfl/chinese-bert-wwm-ext} and BERT-base-uncased\footnote{https://huggingface.co/bert-base-uncased}  as our basic language models for Chinese and English tasks separately. All the classification models are fine-tuned for 5 epochs. The batch size and other information are listed in Table \ref{table6}. Following each epoch of training, we give a test on dev set and record the better model. Finally, the one performs best on dev set will be saved and applied.

\section{Accuracy curves}
\begin{figure*}
    \centering
    \includegraphics[width = 0.8\paperwidth]{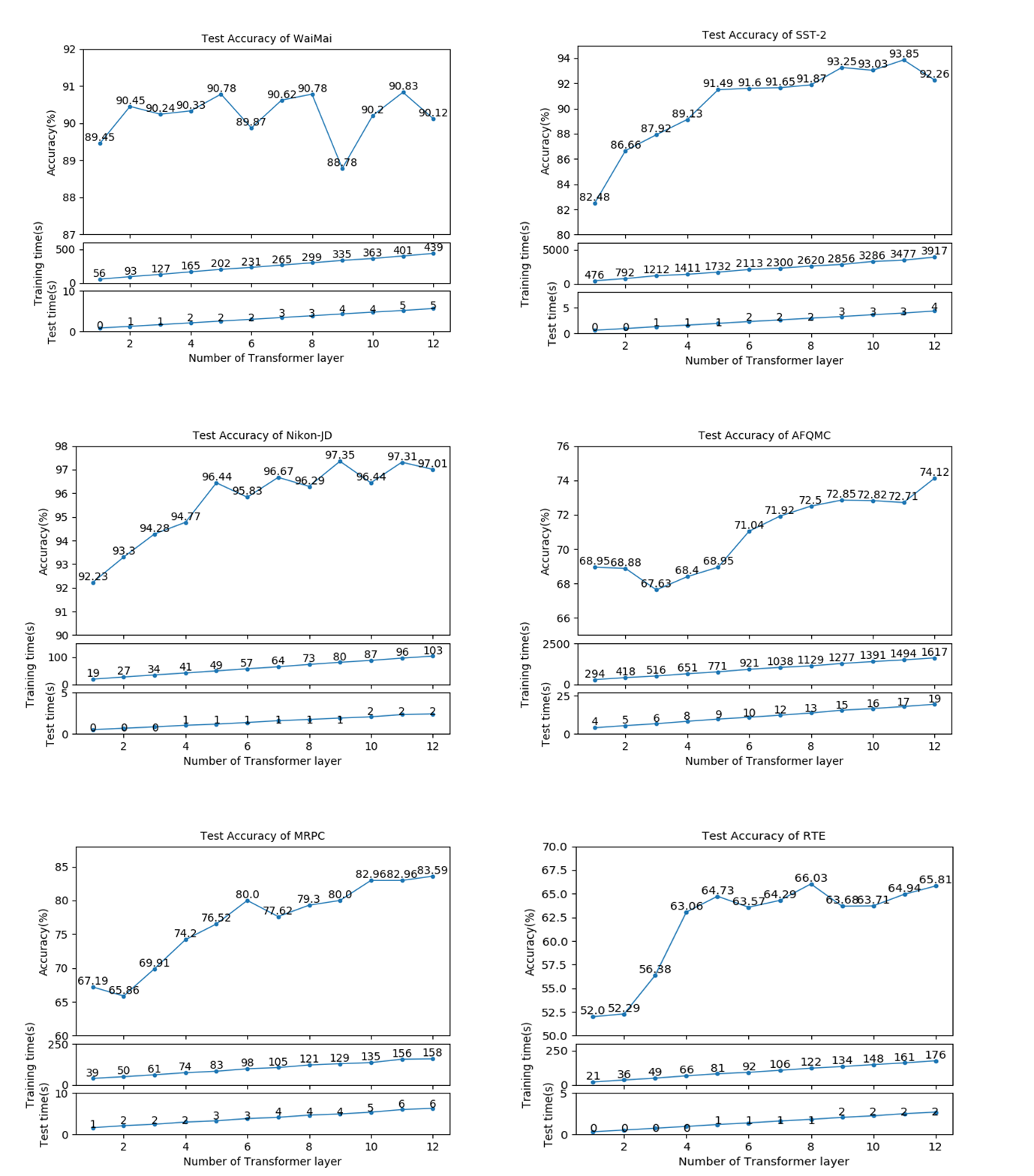}
    \caption{Accuracy, training time and test time of trans1-12. The reservation of accuracy arrives decimally hind two, and training and test time are rounded to the nearest whole number.}
    \label{fig3}
\end{figure*}

In Figure \ref{fig3}, we plot the test accuracy, training time and test time of the other 6 datasets. According to these curves, we confirm that there truly exist some smaller sub-networks surpassing the full BERT model on some downstream tasks. The more difficult the task is, the more complex the corresponding model should be.

\label{apdx-c}

\section{Separability measures and test accuracy}
\label{apdx-d}

In this section, we provide the comparisons of separability measures and test accuracy of other 6 corpora. Similar as the Figure \ref{fig2}, we also capture consistent trends between the CSM measure and test accuracy on these binary classification tasks. That means our truncation method assisted by CSM is resonable and effective.

\begin{figure*}
    \centering
    \includegraphics[width = 0.8\paperwidth]{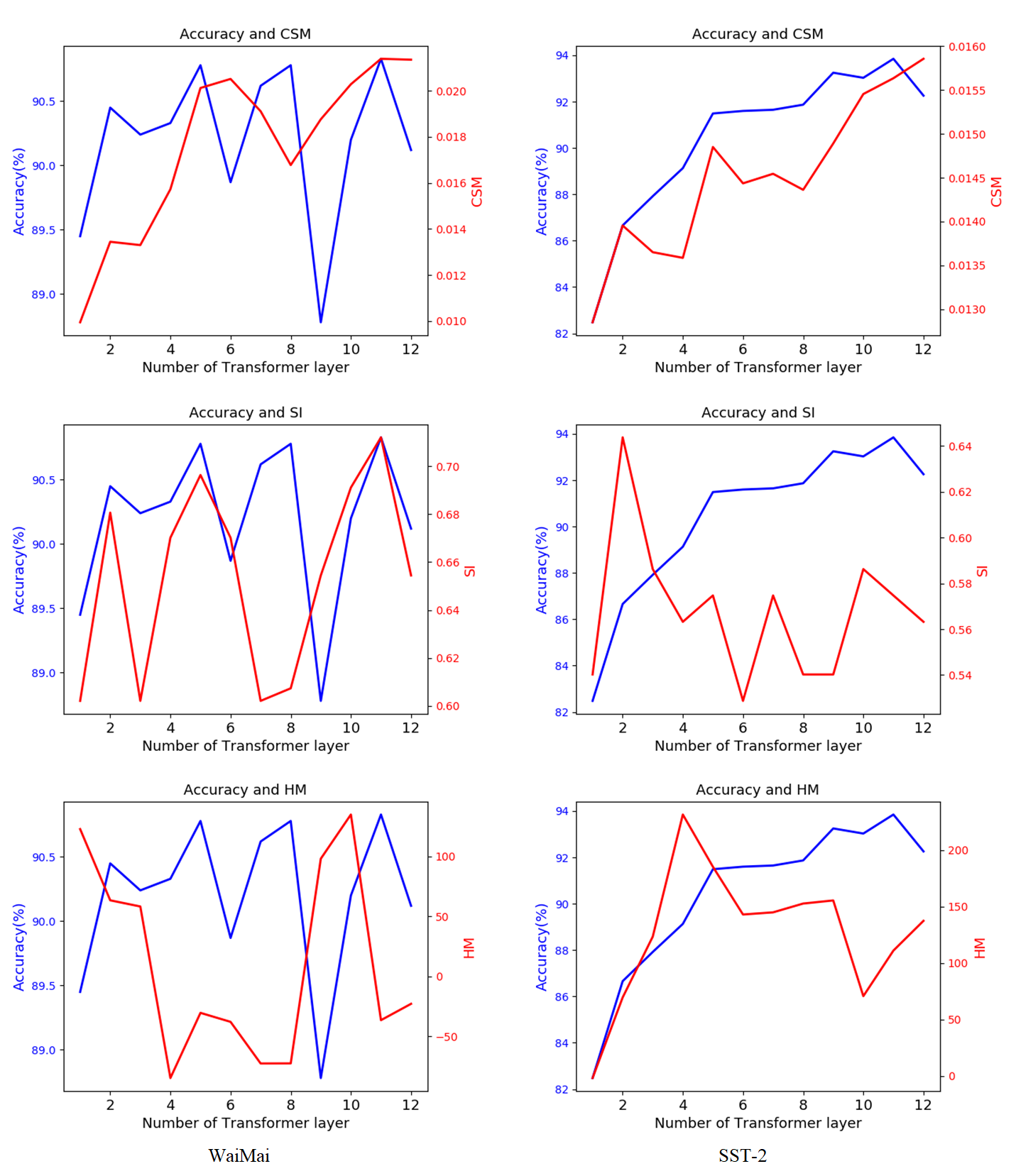}
    \caption{CSM, SI and HM Curves of WaiMai and SST-2. The test accuracy is shown with a royal blue line, and the separability value is drawn with a red line.}
    \label{fig4}
\end{figure*}

\begin{figure*}
    \centering
    \includegraphics[width = 0.8\paperwidth]{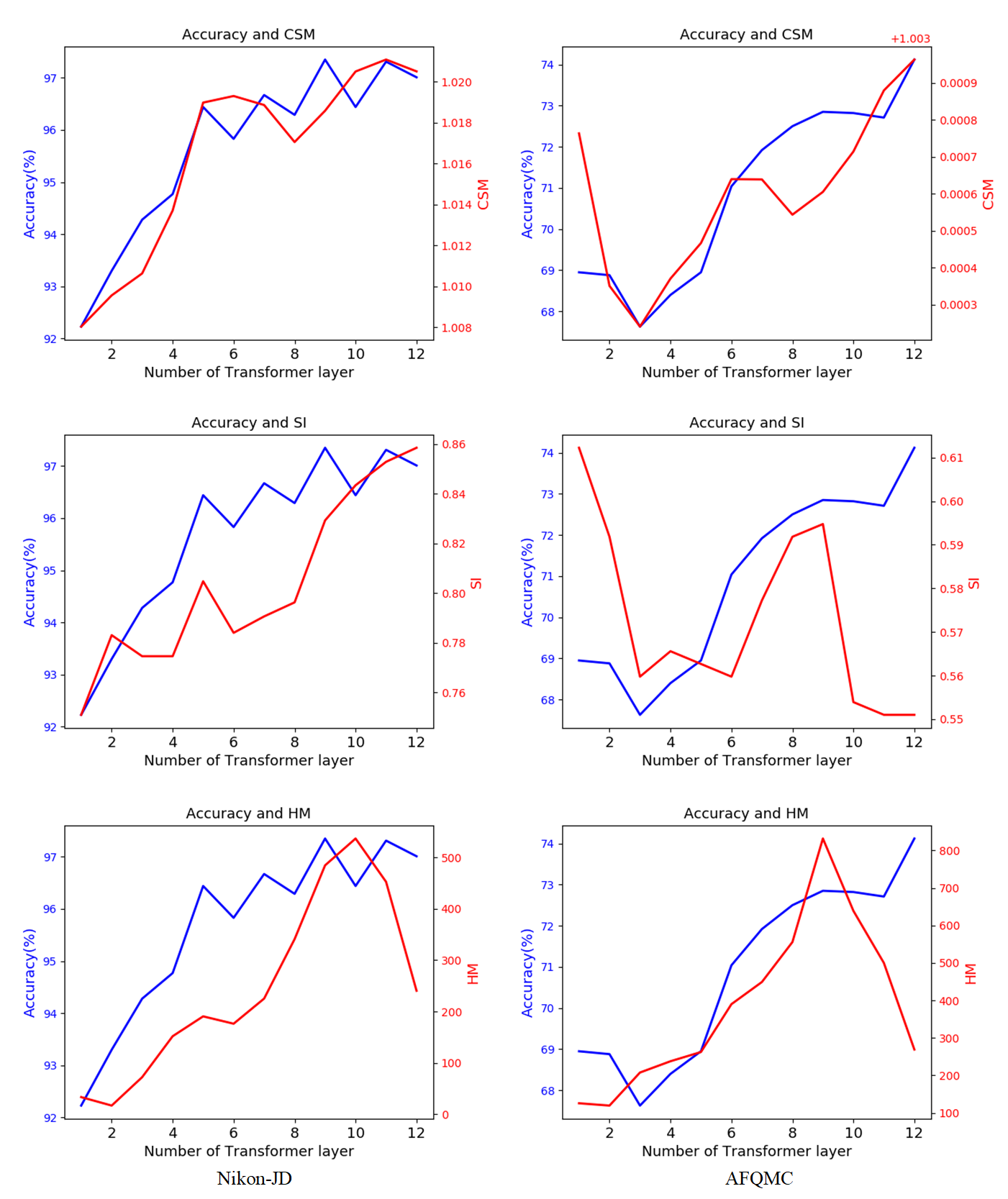}
    \caption{CSM, SI and HM Curves of Nikon-JD and AFQMC. The test accuracy is shown with a royal blue line, and the separability value is drawn with a red line.}
    \label{fig5}
\end{figure*}

\begin{figure*}
    \centering
    \includegraphics[width = 0.8\paperwidth]{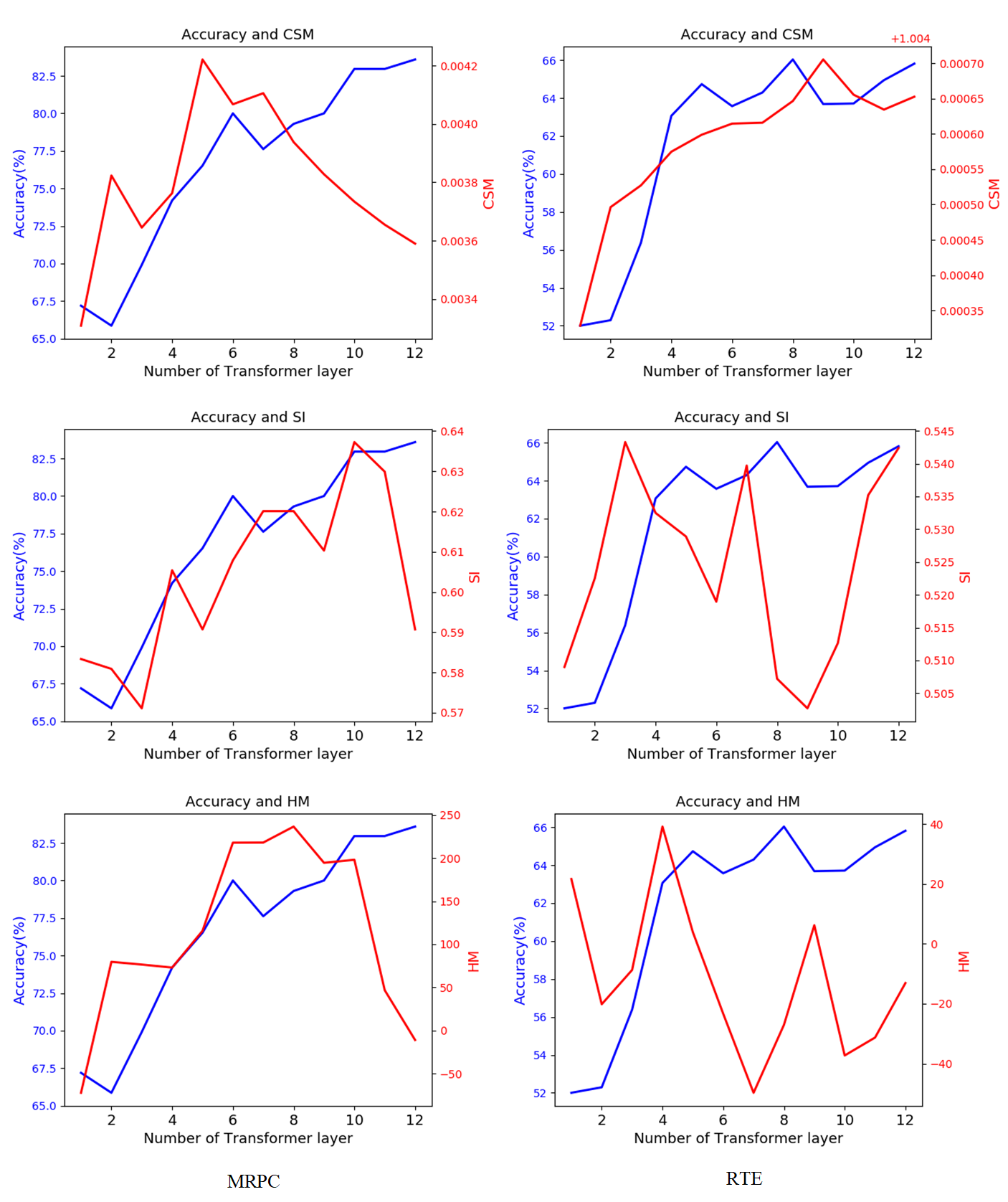}
    \caption{CSM, SI and HM Curves of MRPC and RTE. The test accuracy is shown with a royal blue line, and the separability value is drawn with a red line.}
    \label{fig6}
\end{figure*}

\end{document}